\documentclass[10pt,conference]{IEEEtran}
\IEEEoverridecommandlockouts
\usepackage{cite}
\usepackage{amsmath,amssymb,amsfonts}
\usepackage{arydshln}
\usepackage{algorithmic}
\usepackage{graphicx}
\usepackage{textcomp}
\usepackage{graphicx}
\usepackage{xcolor}
\usepackage{float}
\usepackage{booktabs}
\usepackage{multirow}
\usepackage{soul}
\usepackage{colortbl}
\definecolor{G}{RGB}{144,238,144}
\definecolor{R}{RGB}{255,160,122}
\usepackage{wasysym}
\usepackage{url}

\def\BibTeX{{\rm B\kern-.05em{\sc i\kern-.025em b}\kern-.08em
    T\kern-.1667em\lower.7ex\hbox{E}\kern-.125emX}}
\begin{document}
\title{Unveil Multi-Picture Descriptions for Multilingual Mild Cognitive Impairment Detection via Contrastive Learning}

\author{
    \IEEEauthorblockN{Kristin Qi\textsuperscript{1*}
    , Jiali Cheng\textsuperscript{2*}, Youxiang Zhu\textsuperscript{1}, Hadi Amiri\textsuperscript{2}, Xiaohui Liang\textsuperscript{1}}
    
    \IEEEauthorblockA{\textsuperscript{1}Computer Science, University of Massachusetts, Boston, MA, USA}
    \IEEEauthorblockA{\textsuperscript{2}Computer Science, University of Massachusetts, Lowell, MA, USA}
    
    \IEEEauthorblockA{Email: \{yanankristin.qi001, Youxiang.Zhu001, xiaohui.liang\}@umb.edu, \{jiali\_cheng, hadi\_amiri\}@uml.edu}
}

\maketitle

\begin{abstract}
Detecting Mild Cognitive Impairment from picture descriptions is critical yet challenging, especially in multilingual and multiple picture settings. Prior work has primarily focused on English speakers describing a single picture (e.g., the `Cookie Theft'). The TAUKDIAL-2024 challenge expands this scope by introducing multilingual speakers and multiple pictures, which presents new challenges in analyzing picture-dependent content. 
To address these challenges, we propose a framework with three components: 
(1) enhancing discriminative representation learning via supervised contrastive learning, 
(2) involving image modality rather than relying solely on speech and text modalities, and 
(3) applying a Product of Experts (PoE) strategy to mitigate spurious correlations and overfitting. 
Our framework improves MCI detection performance, achieving a +7.1\% increase in Unweighted Average Recall (UAR) (from 68.1\% to 75.2\%) and a +2.9\% increase in F1 score (from 80.6\% to 83.5\%) compared to the text unimodal baseline. Notably, the contrastive learning component yields greater gains for the text modality compared to speech. These results highlight our framework’s effectiveness in multilingual and multi-picture MCI detection.


\end{abstract}


\section{Introduction}
Mild Cognitive Impairment (MCI) is an intermediate stage between normal cognition and dementia. Early detection of MCI is crucial, as timely interventions before the onset of dementia symptoms can potentially delay disease progression. With the rapid growth of AI applications on the Internet of Things (IoT), these technologies make it possible to detect and monitor cognitive health biomarkers \cite{marshall2015harvard, yamada2021tablet,qi2024exploiting}. 

Traditional cognitive assessments rely on blood tests, neuroimaging, and in-clinic tests like the Mini-Mental State Examination (MMSE). While effective, these methods are time-consuming and labor-intensive. In contrast, AI-driven speech processing has emerged as a tool for early MCI detection through non-invasive and cost-effective screening. These processing systems can automatically identify biomarkers related to cognitive decline, such as hesitation gaps, reduced fluency, vocal tremors, and changes in word richness \cite{beltrami2018speech,peplinski2019objective}.
 
The majority of previous research in dementia detection through speech focused on English-speaking cohorts who describe a single ``Cookie Theft'' picture in the Pitt Corpus \cite{becker1994natural}. Recently, the TAUKDIAL-2024 challenge includes English and Chinese to detect MCI through multilingual speech processing: each participant describes three different pictures in their respective languages \cite{luz2024connected}. However, this multilingual and multi-picture context presents new challenges that reduce the effectiveness of existing models developed for the ``Cookie Theft'' picture description task.

Specifically, TAUKDIAL-2024 introduces two major challenges. First, while studies processing linguistic features in ``Cookie Theft'' speech achieve 85.4\% accuracy \cite{chen2021automatic}, performance on TAUKDIAL-2024 drops to 59.2\% Unweighted Average Recall (UAR) \cite{luz2024connected, gosztolya2024combining, ortiz2024cognitive}. This noticeable drop is due to increased variability between different pictures, which reduces accuracy using linguistic features. Previous approaches do not address complementary information in different pictures and their impact on the detection of cognitive decline.  The second challenge involves the dataset's limited size and imbalanced class labels. Models developed on this speech data risk overfitting and relying on spurious feature correlations rather than true causal ones, which results in shortcut solutions with undermined reliability \cite{cadene2019rubi,liu2024clever,cheng24c_interspeech}.

In this paper, we propose a framework with three components to address the above challenges. For the first challenge, we enhance the model's performance by incorporating picture variability through supervised contrastive learning \cite{khosla2020supervised}. This approach enhances picture-specific feature discrimination by pulling together descriptions of the same picture while pushing apart descriptions of different pictures. We also integrate image embeddings as an additional modality input. Both approaches leverage to improve model abilities with the complementary picture information.  For the second challenge, we adopt a Product of Experts (PoE) strategy for modality concatenation to let multimodalities collectively determine the final prediction label. Since reliance on a single modality leads to performance degradation, PoE prioritizes causal features, and this multimodal decision-making approach mitigates spurious correlations and overfitting \cite{cadene2019rubi,cheng24c_interspeech}. 

In summary, our contributions are threefold:
\begin{itemize}
    \item Our proposed framework leverages contrastive learning to enhance picture-specific discriminative features from multi-picture descriptions. We incorporate image embeddings as an additional modality to improve performance.
    \item  We adopt Product of Expert (PoE) for multimodal processing to mitigate spurious correlations and reduce subgroup disparities, which is crucial for models to generalize using the small-sized and class-imbalanced multilingual speech data. 
    \item  We validate our framework through extensive experiments. Experiments demonstrate a +7.1\% improvement in UAR (68.1\% to 75.2\%) and a +2.9\% improvement in F1 score (80.6\% to 83.3\%) compared to the text unimodal baseline. Notably, our contrastive learning approach enhances performance using text modality compared to speech modality. These results show that our framework effectively addresses the challenges of multilingual, multi-picture MCI detection.  
\end{itemize}  
\begin{figure*}[t]
\centering
\includegraphics[width=\textwidth]{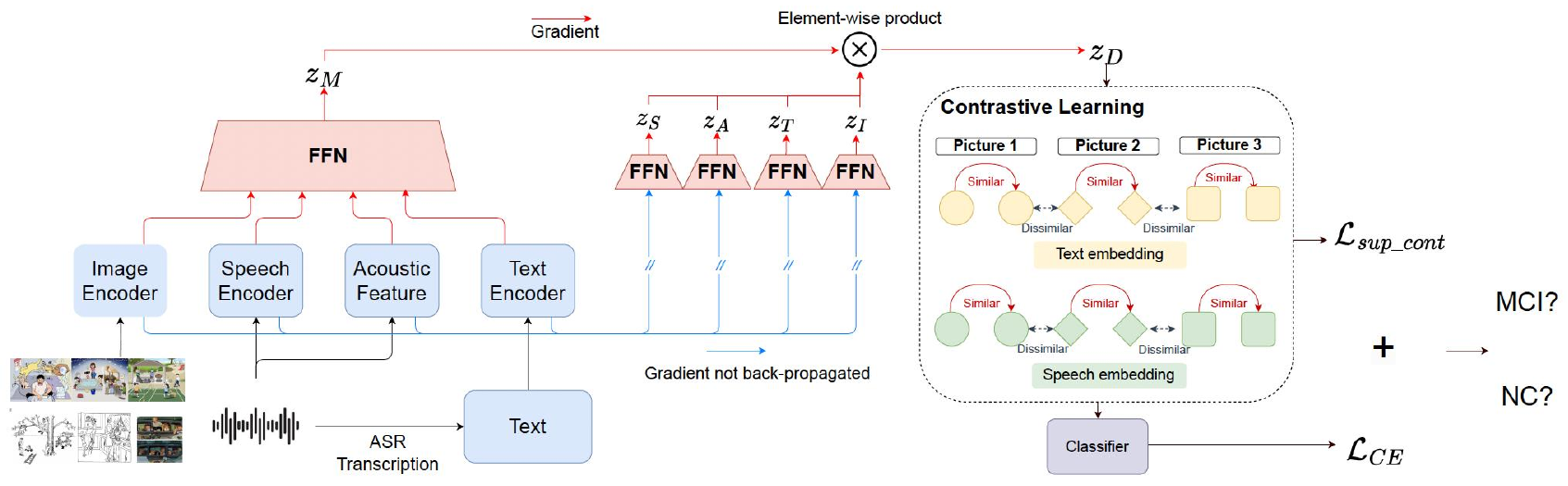}
\caption{Details of components in our framework: (i) the left side shows feature extraction from each unimodality: speech, acoustic, text, and image, and they are processed through feed-forward networks (FFN). (ii) middle shows that PoE fuses these modalities through element-wise product or element-wise addition of their logarithms in log space to produce logits $z_D$. (iii) right side shows supervised contrastive learning to enhance picture-specific feature discrimination. }
\label{fig:architecture}
\end{figure*}

 \section{Related Work}
\subsection{Multilingual Studies in Detecting Cognitive Decline}
Speech and language processing for cognitive decline detection spans several decades, with most methods focusing on English speakers. Two widely used datasets are the ADReSS challenge and Pitt Corpus \cite{becker1994natural}. Speech processing combines acoustic and linguistic features with machine learning classifiers. Studies have reported accuracy ranging from 72.9\% to 89.6\% depending on models \cite{gauder2021alzheimer, balagopalan2021comparing}. A few studies include more languages. ADReSS-M challenge includes English and Greek speakers for dementia detection, achieving accuracy up to 87\% \cite{luz2023multilingual}. Pérez-Toro \emph{et al.} have included English, German, and Spanish speakers for Alzheimer's detection \cite{perez2023automatic} and achieved UAR of 73\%. Furthermore, the TAUKDIAL-2024 challenge focuses on MCI detection, and the established baseline achieves 59.18\% UAR. Other reported UARs are between 49.4\% and 56.18\% \cite{gosztolya2024combining, ortiz2024cognitive}. These results imply the challenges in developing highly performing and generalizable models for multilingual MCI detection.
  
\subsection{Overfitting and Spurious Correlation}
Recent work by Liu \emph{et al.} \cite{liu2024clever} raised important questions about the reliability of reported high accuracy in dementia detection. Their analysis of the Pitt Corpus revealed that models could achieve good results by relying solely on silence segments, which is the Clever Hans effect. This finding suggests that models might be exploiting dataset-specific shortcuts, leading to spurious correlations rather than learning true causal patterns. Spurious correlations can come from imbalanced and noisy datasets. The consequence of spurious correlations is bias against minorities and over-reliance on data shortcuts, limiting real-world application deployment \cite{yang2023change}. While rectifying the weakness of relying on spurious correlation is crucial for clinical applications, it remains a challenge in adapting existing algorithms to medical routines. To address the challenges, we adopt PoE to mitigate reliance on spurious features and reduce performance disparities between subgroups \cite{karimi-mahabadi-etal-2020-end,cheng-amiri-2024-fairflow}.

\section{Dataset}
\label{section:data}
The TAUKADIAL-2024 challenge consists of speech recordings from the participants aged 60 to 90 years old (mean age: 72), with different numbers of male and female speakers. It includes English and Chinese-speaking participants and maintains three recordings for each participant, where each participant describes three different pictures. 

The last digit in each recording file name indicates the specific picture being described, with 1, 2, and 3 corresponding to the first, second, and third pictures, respectively. Six pictures were described, three in English and three in Chinese. We relabeled each picture as 1-6 in our data processing. 

For each participant, a standard clinical assessment using the MMSE was performed. Based on their MMSE scores, participants were labeled as either normal cognitive decline (NC) or MCI. The labels are provided in this dataset. The dataset comprises 129 participants (74 MCI and 55 NC), resulting in a total of 387 samples (222 MCI and 165 NC). Of these samples, 186 are English speakers and 201 are Chinese speakers. Details of the dataset are described in Table \ref{tab:subject_summary}.

We are aware that previous work \cite{luz2024connected, ortiz2024cognitive, hoang2024translingual, gosztolya2024combining} used the same speech data and reported higher metrics under cross-validation settings; however, their performance on the test set remains substantially lower (below 60\%). Since the data does not release the test set publicly, we report
cross-validation performance only and do not compare against
previous work on the test set.
\begin{table}[t]
\caption{Dataset statistics with English and Chinese speakers.}
\centering
\begin{tabular}{lcccccc}
\hline
\textbf{} & \textbf{Age} & \textbf{Male} & \textbf{Female} & \textbf{MMSE} & \textbf{English} & \textbf{Chinese} \\
\hline
\textbf{MCI} & 73.36 & 87 & 134 & 25.84 & 123 & 98 \\
\textbf{NC}  & 71.85 & 63 & 102 & 29.07 & 63  & 102 \\
\hline
\end{tabular}
\label{tab:subject_summary}
\end{table}

\section{Problem Formulation}
Our framework integrates embeddings from speech, text, and picture inputs to classify participants as NC or MCI in multilingual settings. Suppose the dataset is \( D = \{(X_i, Y_i)\}_{i=1}^N \) containing \( N \) speech recording samples. Each sample $X_i = (S_i, T_i, I_i)$ consists of three input modalities: speech ($S_i$), text ($T_i$), and image ($I_i$). The corresponding label $Y_i \in \{0,1\}$ indicates NC or MCI, respectively. The task is to predict $\hat{Y}_i \in \{0,1\}$ for each sample $X_i$.
 
\section{Methods}
Our framework is shown in Figure \ref{fig:architecture}. Embeddings from multimodality are combined into a joint representation. This representation is processed through the PoE and trained with supervised contrastive loss and cross-entropy loss to produce final labels. In this section, we explain the details of feature extraction, contrastive learning, and the PoE approach.

\subsection{Preprocessing}
For preprocessing speech inputs, we first removed the noisy background and applied a 16 kHz sampling rate to the audio with a single (mono) channel. Speech was transcribed using the Whisper large-v3 \cite{radford2023robust} 
automatic speech recognition (ASR) model to convert audio recordings into text. We maintained the default feature dimension settings for pre-trained models.
\subsection{Feature Extraction}
We use established models for extracting features from speech and transcriptions.
\label{section:feature}

\noindent\textbf{Speech Encoder:}  We process each speech input $S_i$ through a speech encoder $f_S$ to obtain speech embeddings $f_S(S_i)$. Specifically, we use the pre-trained Whisper-tiny model 
\cite{radford2023robust}, which extracts features from a transformer-based architecture.

\noindent\textbf{Acoustic Features:}   The handcrafted acoustic features supplement previous speech encoder features. We extract a comprehensive set of acoustic features $f_A(S_i)$ using the DisVoice 
package \cite{martinez2021ten}. These handcrafted features are phonation characteristics, phonological information, articulation dynamics, and prosodic elements, which are relevant to identifying the MCI speech patterns.

\noindent\textbf{Text Encoder:}    For each transcription $T_i$, linguistic embeddings $f_T(T_i)$ are obtained using the pre-trained language model. We use the BERT-base
\cite{devlin-etal-2019-bert} as the text encoder. 

\noindent\textbf{Image Encoder:} For each audio recording where a participant describes a picture, we obtain image embeddings from the corresponding picture using ResNet-34 \cite{he2016deep} as the image encoder $f_I$. Specifically, the encoder maps each picture being described $I_i$ to its embedding representation $f_I(I_i)$.

\subsection{Contrastive Learning for Feature Enhancement}
\label{section:text} 
To enhance discrimination between descriptions of different pictures, we employ contrastive learning. This approach learns to pull together positive pairs (similar descriptions) while pushing apart negative pairs (dissimilar descriptions). Ultimately, it aims to minimize distances between positive pairs while maximizing distances between negative pairs. 

We adopt supervised contrastive learning \cite{khosla2020supervised} where pair relationships are determined by picture labels (see Section \ref{section:data}). We define positive pairs as descriptions of the same picture and negative pairs as descriptions of different pictures. Since positive pairs have similar embeddings and negative pairs have distant embeddings, the model learns to minimize distances between positive pair embeddings while maximizing distances between negative pair embeddings. Thus, contrastive learning enforces the learning of both picture-specific features and cognitive status features (NC vs. MCI). The equation of contrastive learning loss ($\mathcal{L}_{\text{sup\_cont}}$) is shown below:

\begin{equation}
\mathcal{L}_{\text{sup\_cont}} = \sum_{k \in \mathcal{B}} \frac{-1}{|P(k)|} \sum_{p \in P(k)} \log \frac{\exp\left(\frac{\mathbf{h}_k \cdot \mathbf{h}_p}{\tau}\right)}{\sum_{n \in N(k)} \exp\left(\frac{\mathbf{h}_k \cdot \mathbf{h}_n}{\tau}\right)},
\end{equation} where $\mathbf{h}_k$ denotes the embedding of a description in the batch. $\mathcal{B}$ represents the set of batches. $P(k)$ is the set of descriptions of the same picture (positive) as indices $k$. $N(k)$ is the set of descriptions of different pictures (negative). $\tau$ is a temperature parameter controlling the scaling of embedding similarity scores.

\subsection{Product of Experts for Multimodal Concatenation}
We adopt the Product of Experts (PoE) approach rather than simple concatenation of modalities, which can rely on spurious features and shortcut solutions \cite{cadene2019rubi}.  PoE model learns from weaker modalities and balances individual contributions, preventing stronger modalities from dominating the prediction. We begin by processing the features extracted from each unimodality (see Section \ref{section:feature}). First, we concatenate these features from unimodalities and transform them through feed-forward layers (FFN):
\vspace{-1pt}
\begin{equation}
z_M = \mathrm{FFN}_M([f_S(S_i); f_A(S_i); f_T(T_i); f_I(I_i)])
\end{equation}
where $[;]$ denotes concatenations. $\mathrm{FFN}_M$ is a feed-forward network to transform multimodal concatenation. $z_M$ denotes the predicted class logits from the multimodal concatenation.  

For each unimodality, we also obtain individual predictions of class labels through separate feed-forward networks:
\begin{equation}
    z_S = \mathrm{FFN}_S(f_S(S_i)).
\end{equation}
where $\mathrm{FFN}$ is a modality-specific feed-forward network that maps corresponding features to class logits. Similarly, we do this for all other unimodalities.

Next, we aim to combine these predictions efficiently, where the PoE operates in the log space. Rather than performing element-wise multiplication of probabilities, we sum their logarithm values (element-wise addition) because log-space operation is more efficient than probability multiplication.
\begin{equation}
\log z_D = \log z_S + \log z_A + \log z_T + \log z_I.
\end{equation}
where $z_D$ denotes the final decision logits.

This combination ensures that unimodalities collectively contribute to the final predictions, preventing the reliance on spurious features or a dominated modality. $z_D$ represents a more robust prediction than multimodal concatenation, since it implements dynamic sample weighting: each modality's contribution is adaptively adjusted based on its confidence for each input example. The final prediction logits are used to compute the cross-entropy (CE) loss for binary classification between NC and MCI. 

\subsection{Training} 
The total loss function is a combination of binary cross-entropy loss ($\mathcal{L}_{\text{CE}}$) and the supervised contrastive learning loss ($\mathcal{L}_{\text{sup\_cont}}$), where $\lambda =1$ balances the two component weights:
\begin{equation}
\mathcal{L}_{\text{total}} = \mathcal{L}_{\text{CE}} + \lambda \mathcal{L}_{\text{sup\_cont}}
\end{equation}

\section{Experiments}
\label{sec:experimental_setup}      
\subsection{Baseline Details}
\label{section:baseline}
These baselines focus on each unimodality: $z_S$ represents speech features extracted from the pre-trained speech encoder. $z_A$ represents handcrafted acoustic features from the feature extractor. $z_T$ represents linguistic features from the text encoder. We include $z_M = z_S+ z_A + z_T$ to represent speech and text modality concatenation.
 
\subsection{Implementation Details}
\label{sec:detail}
To ensure robust evaluation given small and imbalanced data, we trained our framework with stratified k-fold cross-validation (k=10). Hyperparameters include a learning rate of 1e-5, a batch size of 16, L2 regularization penalty$=0.01$, and the Adam optimizer. The framework is trained for 10 epochs and implemented in PyTorch on an A100 GPU.

\subsection{Evaluation Metrics}
We adhere to evaluation metrics required by the challenge for binary classification: UAR and F1 score: \(\text{UAR} = \frac{\sigma + \rho}{2}\) and \(\text{F1} = \frac{2\pi\rho}{\pi + \rho}\), where \(\sigma = \frac{\text{TN}}{\text{TN} + \text{FP}}\) represents specificity, \(\rho = \frac{\text{TP}}{\text{TP} + \text{FN}}\) represents sensitivity, and \(\pi = \frac{\text{TP}}{\text{TP} + \text{FP}}\) represents precision. \(\text{TP}\), \(\text{TN}\), \(\text{FP}\), and \(\text{FN}\) denote true positives, true negatives, false positives, and false negatives, respectively.

\section{Results and Discussion}
We focus on the effect of adding supervised contrastive learning (CL), image embeddings (IE), and PoE into baseline models, which are discussed in section~\ref{section:baseline}. We evaluate our framework on (1) the entire dataset (``Both'') and further break down the dataset into (2) the language subgroup: English (``En'') and Chinese (``Zh''), and (3) the gender subgroup: Female (``F'') and Male (``M''). It is particularly noted that $z_A$ represents the handcrafted features and is non-learnable with contrastive learning.

\subsection{Baseline Performance}
The baseline details are discussed in \ref{section:baseline}. Shown in Table \ref{tab:proposed_method_comparison}, the baseline models use unimodal features. On the combined dataset ``Both'', speech encoder ($z_S$) achieves UAR of 69.9\%. Acoustic features ($z_A$) achieve UAR of 68.2\%, and the text encoder ($z_T$) achieves UAR of 68.1\%. These results indicate that relying on individual modalities is insufficient for achieving highly accurate MCI detection in the multilingual setting. Regarding subgroups of language and gender, Chinese performance is generally higher than English, and male group is better than the female group.
\begin{table}[h]
\caption{Results of unimodality with CL and IE, for entire dataset and each language and gender subgroup. Green values indicate performance gain over baseline.}
\label{tab:proposed_method_comparison}
\centering
\renewcommand{\arraystretch}{1.2} 
\Huge
\resizebox{\columnwidth}{!}{%
\begin{tabular}{l|ccccc|ccccc}
\toprule
\textbf{Method} & \multicolumn{5}{c|}{\textbf{UAR}} & \multicolumn{5}{c}{\textbf{F1}}\\
\cmidrule(lr){2-6}\cmidrule(lr){7-11}
& \textbf{Both} & \textbf{En} & \textbf{Zh} & \textbf{M} & \textbf{F}
& \textbf{Both} & \textbf{En} & \textbf{Zh} & \textbf{M} & \textbf{F} \\
\midrule
\hline
\textbf{$z_S$}     
& 69.9 & 55.9 & 74.2 & 80.3 & 63.3 & 80.6 & 79.6 & 75.7 & 84.3 & 77.5 \\
\textbf{+ CL}      
& 68.0 & 55.0 & \cellcolor{G}74.5 & 75.5 & 62.9 & \cellcolor{G}80.7 & 79.2 & \cellcolor{G}78.6 & 82.3 & \cellcolor{G}78.5 \\
\textbf{+ CL + IE} 
& 69.0 & 55.0 & 72.7 & 78.8 & 63.0 & 80.2 & 79.2 & 74.3 & 83.3 & 77.2 \\
\midrule
\textbf{$z_A$}
& 68.2 & 55.0 & 70.3 & 77.3 & 61.6 & 78.9 & 79.2 & 71.8 & 79.2 & 75.6 \\
\textbf{+ IE}       
& \cellcolor{G}68.3 & 49.5 & \cellcolor{G}70.4 & \cellcolor{G}78.9 & 61.3 & 78.7 & 78.7 & 71.6 & \cellcolor{G}79.7 & 75.1 \\
\midrule
\textbf{$z_T$}
& 68.1 & 55.0 & 74.7 & 76.1 & 62.5 & 80.6 & 79.2 & 79.0 & 82.8 & 78.1 \\
\textbf{+ CL}      
& \cellcolor{G}71.8 & \cellcolor{G}60.9 & 74.6 & 74.2 & \cellcolor{G}71.3 & \cellcolor{G}81.5 & \cellcolor{G}80.4 & 72.8 & 81.1 & \cellcolor{G}81.3 \\
\textbf{+ CL + IE} 
& \cellcolor{G}73.1 & \cellcolor{G}61.6 & 74.1 & \cellcolor{G}76.5  & \cellcolor{G}71.8 & \cellcolor{G}82.0 & \cellcolor{G}81.1 & 71.6 & 82.6 & \cellcolor{G}81.0 \\
\midrule
\textbf{$z_M$}
& 72.4 & 58.0 & 83.4 & 83.1 & 70.0 & 81.3 & 80.3 & 78.7 & 87.0 & 78.4 \\
\textbf{+ CL}      
& \cellcolor{G}72.6 & 56.1 & 73.6 & 78.0 & 67.9 & \cellcolor{G}82.2 & \cellcolor{G}81.5 & 74.3 & 86.8 & \cellcolor{G}79.5 \\
\textbf{+ CL + IE} 
& \cellcolor{G}73.5 & 55.1 & 77.3 & 82.5 & 68.3 & \cellcolor{G}83.3 & \cellcolor{G}81.0 & 77.4 & 85.8 & \cellcolor{G}79.6 \\
\bottomrule
\end{tabular}
}
\end{table}
\vspace{-5pt}

\subsection{After Adding CL and IE to Unimodality}
\noindent\textbf{CL and IE are more effective for the text modality than for the speech modality.   } As shown in Table \ref{tab:proposed_method_comparison}, adding CL to $z_T$ increases UAR from 68.1\% baseline to 71.8\% (+3.7\%). The addition of IE further improves it by 5.0\% (73.0\%). F1 score improves from 80.6\% to 82.0\% after adding both CL and IE. In contrast, speech models show no substantial improvements. These indicate that CL can effectively enhance the model's ability to discriminate linguistic features and semantic differences from multi-picture descriptions. The inclusion of IE provides more picture context in addition to text. 

\noindent\textbf{Language and gender subgroups with unimodality.    } In terms of language-specific performance, CL and IE demonstrate more improvements in English compared to Chinese within the text modality: adding CL and IE noticeably increases the UAR in English from 55.0\% to 61.6\% (+6.6\%). In terms of gender subgroup performance, improvements are consistent, where the text modality with both CL and IE improves the UAR of the female by +9.3\% (62.5\% to 71.8\%).

\subsection{After Combining All Modalities}
\noindent\textbf{CE and IE integration with PoE improves multimodal baseline performance.   } Shown in Table \ref{tab:multi}, our multimodal baseline $Z_M$ achieves UAR of 72.4\%, outperforming all unimodal approaches. Incorporating PoE further improves UAR to 73.6\%, since PoE encourages individual modalities to contribute collectively to make the final decision. This process aims to enhance model generalization. When we combine PoE, CL, and IE, we observe additional improvements compared with $Z_M$: increases in UAR by +2.8\% (72.4\% to 75.2\%) and F1 score by +2.2\% (81.3\% to 83.5\%). These results imply that the combination of CL and IE with PoE leads to an improved framework for MCI detection.
\begin{table}[h]
\caption{Comparison of multimodal baseline ($z_M$), with PoE, CL and IE. Bold values indicate $z_M$ + PoE outperforms $z_M$. Green highlights indicate improvements over $z_M$ + PoE.}
\label{tab:multi}
\centering
\renewcommand{\arraystretch}{1.3} 
\Huge 
\resizebox{\columnwidth}{!}{%
\begin{tabular}{l|ccccc|ccccc}
\toprule
\textbf{Method} & \multicolumn{5}{c|}{\textbf{UAR}} & \multicolumn{5}{c}{\textbf{F1}}\\
\cmidrule(lr){2-6}\cmidrule(lr){7-11}
& \textbf{Both} & \textbf{En} & \textbf{Zh} & \textbf{M} & \textbf{F}
& \textbf{Both} & \textbf{En} & \textbf{Zh} & \textbf{M} & \textbf{F} \\
\midrule
\hline
\textbf{$z_M$}     
& 72.4 & 58.0 & 83.4 & 83.1 & 70.0 & 81.3 & 80.3 & 78.7 & 87.0 & 78.4 \\
\midrule
\textbf{$z_M$ + PoE}      
& \textbf{73.6} & \textbf{58.9} & 78.7 & 78.5 & \textbf{73.2} & \textbf{81.7} & \textbf{80.6} & \textbf{79.5} & 80.9 & \textbf{82.1} \\
\textbf{+ CL} 
& \cellcolor{G}73.8 & \cellcolor{G}59.0 & 78.7 & \cellcolor{G}78.8 & 73.2 & \cellcolor{G}82.3 & \cellcolor{G}81.3 & \cellcolor{G}79.8 & \cellcolor{G}81.3 & 82.0 \\
\textbf{+ CL + IE} 
& \cellcolor{G}75.2 & \cellcolor{G}60.1 & 78.3 & \cellcolor{G}78.6 & 69.3 & \cellcolor{G}83.5 & 80.6 & \cellcolor{G}79.7 & \cellcolor{G}84.4 & 80.2 \\
\bottomrule
\end{tabular}}
\end{table}

\noindent\textbf{Language and gender subgroup with multimodality.  } Language-specific results show that PoE improves English performance, with +0.9\% increase in UAR and with a further increase of +2.1\% (58.0\% to 60.1\%) after adding CL and IE. While Chinese shows stronger baseline performance, its UAR decreases after adding PoE. This pattern suggests that PoE balances the integration of modalities across languages, reducing inter-group disparities. Similar results emerge across gender subgroups: the lower-performing female baseline shows improvements of +3.2\% in UAR (70.0\% to 73.2\%) and +3.7\% in F1 score (78.4\% to 82.1\%), while male performance drops, narrowing the gender disparity. Thus, we fairly argue that PoE has the capability to balance learning from different modalities regardless of differences between subgroups for a better generalization. The addition of CL and IE further improves the male F1 score by +3.5\% (80.9\% to 84.4\%), though the female performance remains stable. 

\noindent\textbf{PoE reduces disparity across subgroups.    } The inclusion of PoE leads to improvements either in UAR or F1 score compared to underperforming baseline subgroups. In the gender subgroup, female has UAR increase of +3.2\% (70.0\% to 73.2\%) and F1 score increase of +3.7\% (78.4\% to 82.1\%). The male performance significantly reduces, thereby reducing the gender disparity. Similarly, since Chinese UAR reduces while English remains stable, the gap between the previously stronger Chinese baseline narrows. These results exhibit that PoE effectively elevates worse-performing subgroups while maintaining outcomes from better-performing subgroups.

\subsection{The Impact of Contrastive Learning on Other Models}
To further understand the effectiveness of adding CL, we evaluate it across multiple linguistic and speech feature extraction models. Linguistic models include 
BERT, 
RoBERTa\footnote{\url{https://huggingface.co/FacebookAI/roberta-base}} 
\cite{liu2019roberta}, and XLM\footnote{\url{https://huggingface.co/FacebookAI/xlm-roberta-base}} 
\cite{conneau2019cross}. Speech models include Whisper, HuBERT\footnote{\url{https://huggingface.co/facebook/hubert-base-ls960}} 
\cite{hsu2021hubert}, and Wav2Vec2\footnote{\url{https://huggingface.co/facebook/wav2vec2-base}} 
\cite{baevski2020wav2vec}. 
\begin{figure}[t]
    \centering
    \includegraphics[width=0.46\textwidth]{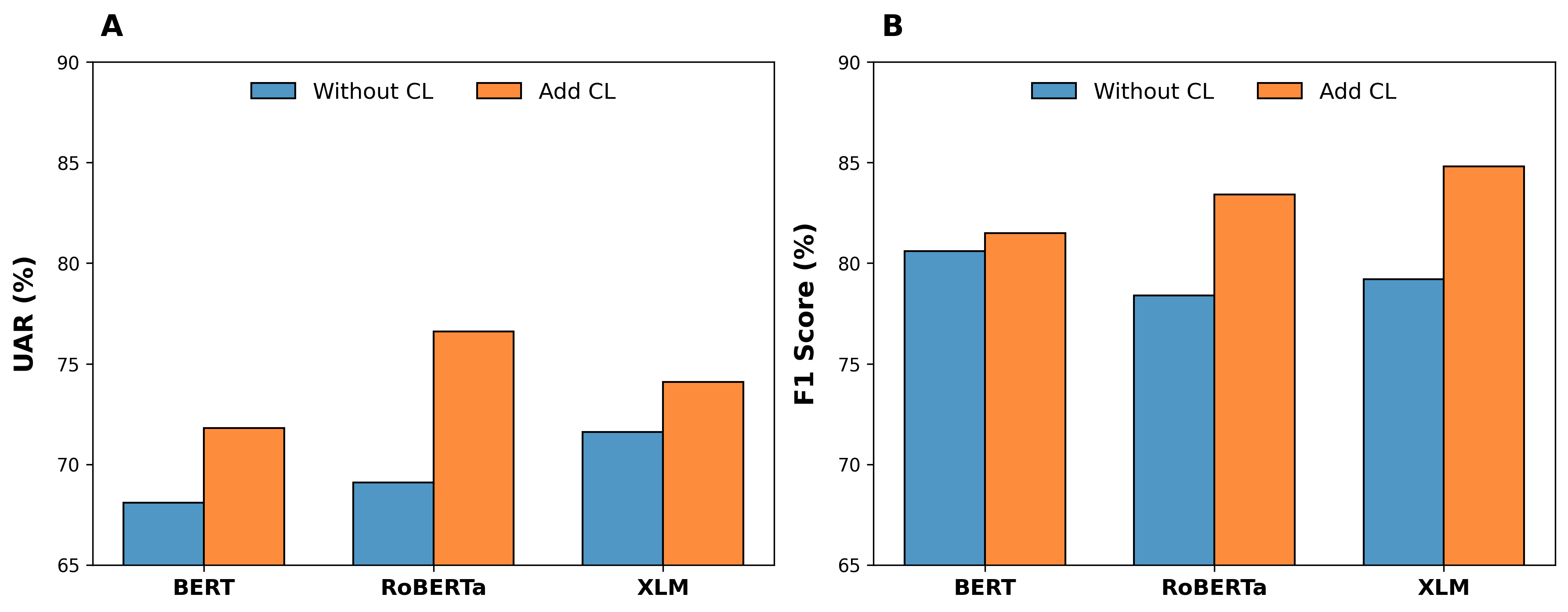}
    \includegraphics[width=0.46\textwidth]{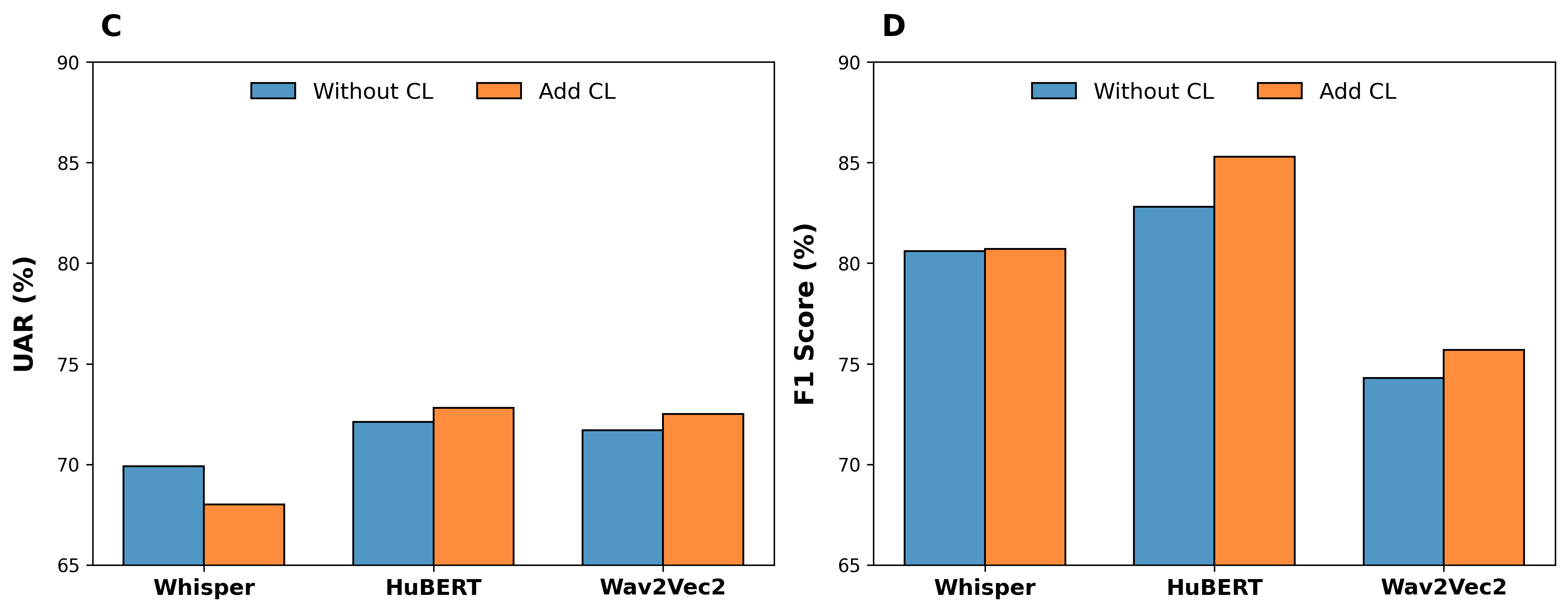}
    \caption{Performance impact of contrastive learning on different models. \textbf{A-B:} UAR and F1 scores for linguistic models (BERT, RoBERTa, XLM) with and without CL. \textbf{C-D:} UAR and F1 scores for speech models (Whisper, HuBERT, Wav2Vec2) with and without CL.}
    \label{fig:model_comparison}
\end{figure}

\noindent\textbf{Linguistic models reveal consistent improvements with CL.  } As shown in Figures \ref{fig:model_comparison}A and \ref{fig:model_comparison}B, RoBERTa shows a noticeable improvement, with UAR increasing by +7.5\% and F1 score by +5.2\%. XLM has consistent improvements (UAR +2.5\%, F1 +5.6\%). Notably, XLM has a stronger UAR baseline prior to adding CL, likely due to its cross-lingual pre-training. In contrast, the effect of CL on speech features demonstrates limited benefits in Figures \ref{fig:model_comparison}C and \ref{fig:model_comparison}D: HuBERT shows a small increase of +0.7\% in UAR and +2.5\% in F1 score, as well as Wav2Vec2 (UAR +0.8\%, F1 score +1.4\%). HuBERT has a strong baseline among speech models.

\noindent\textbf{Contrastive learning enhances picture-specific linguistic features more effectively.   } Our experiments consistently demonstrate that CL better separates linguistic feature embeddings than speech features in high-dimensional spaces. This enhanced separation improves discrimination between descriptions of different pictures in the linguistic domain. However, speech embeddings prove less effective for this separation because they are dominated by speaker-specific characteristics (vocal patterns, speaking style) rather than picture-specific content. Consequently, picture-specific differences remain more distinguishable in linguistic representations than in speech feature space.


\section{LIMITATION}  Our work primarily uses the supervised contrastive learning loss and PoE loss functions. We hypothesize that the results will generalize beyond the loss functions we discussed. Future validation is needed to verify this. Other questions remain open: we explore a limited set of MCI-relevant handcrafted features. Future work will analyze more handcrafted features and how they are correlated with the performance of our framework. We will also compare with other multimodal fusion strategies to compare their performance with ours. In addition, the small and imbalanced data limit the highly performing features and models, which require future data collection and benchmark efforts.

\section{Conclusion}
In this paper, we present a framework for MCI detection that addresses the challenges of multilingual and multi-picture analysis. By integrating supervised contrastive learning, image embeddings, and a Product of Experts (PoE) mechanism, our approach enhances MCI-related feature representations while mitigating modality-specific biases. The proposed method achieves notable improvements over the text-only baseline, with a +7.1\% gain in UAR (from 68.1\% to 75.2\%) and a +2.9\% gain in F1 score (from 80.6\% to 83.5\%). Our findings demonstrate that contrastive learning effectively captures picture-specific discriminative features, particularly in the text modality. Furthermore, PoE-based multimodal fusion balances modality contributions and improves generalization by reducing spurious correlations and subgroup disparities. \\

\section{Acknowledgement}  
This research is funded by the US National Institutes of Health National Institute on Aging R01AG067416.  

\bibliographystyle{IEEEtran}
\bibliography{mybib}

\end{document}